\documentclass[11pt]{article}

\usepackage[utf8]{inputenc}
\usepackage[T1]{fontenc}
\usepackage{hyperref}
\usepackage{url}
\usepackage{booktabs}
\usepackage{amsmath}
\usepackage{graphicx}
\usepackage{float}
\usepackage[margin=0.9in]{geometry}
\usepackage{natbib}

\hypersetup{
    colorlinks=true,
    linkcolor=blue,
    filecolor=magenta,
    urlcolor=blue,
    citecolor=blue
}

\title{Multi-Agent Reinforcement Learning for Cooperative Warehouse Automation: QMIX Value Decomposition for Sparse-Reward Coordination}

\author{Price Allman, Lian Thang, Dre Simmons, Salmon Riaz \\
\textit{Course: Artificial Intelligence,}\\
\textit{Department of Computer Science, Oral Roberts University, Tulsa, OK, USA}}

\date{December 2025}

\begin{document}

\maketitle

\begin{center}
\textbf{Preamble}\\
This paper was written as part of the Artificial Intelligence course at Oral Roberts University. Its purpose is educational: to document the methodologies, analyses, and results of the project completed during the Fall semester. Although it follows academic standards, it is not a formal research publication.
\end{center}

\begin{abstract}
We present a comparative study of multi-agent reinforcement learning (MARL) algorithms for cooperative warehouse robotics. We evaluate QMIX and IPPO on the Robotic Warehouse (RWARE) environment and a custom Unity 3D simulation. Our experiments reveal that QMIX's value decomposition significantly outperforms independent learning approaches (achieving 3.25 mean return vs. 0.38 for advanced IPPO), but requires extensive hyperparameter tuning---particularly extended epsilon annealing (5M+ steps) for sparse reward discovery. We demonstrate successful deployment in Unity ML-Agents, achieving consistent package delivery after 1M training steps. While MARL shows promise for small-scale deployments (2-4 robots), significant scaling challenges remain. Code and analyses: \href{https://pallman14.github.io/MARL-QMIX-Warehouse-Robots/}{project documentation}.
\end{abstract}

\section{Introduction}

Our research investigates the complex task of multiple autonomous agents learning to coordinate and deliver packages in warehouse environments---a problem requiring implicit communication, collision avoidance, and efficient task allocation without centralized control. Traditional warehouse automation relies on centralized planning systems that face scalability limitations; multi-agent reinforcement learning (MARL) offers an alternative through decentralized learned policies, but requires solving the credit assignment problem.

We compare MARL algorithms on warehouse coordination: QMIX \citep{rashid2018qmix} (value decomposition), IPPO (independent learning), and MASAC (centralized critic). Our study progresses from MPE for validation to RWARE for warehouse evaluation, culminating in Unity 3D deployment where agents demonstrate learned package delivery behavior. QMIX emerged as the best performer after systematic comparison.

Our contributions: (1) hyperparameter analysis showing default configurations fail on sparse-reward warehouse tasks, (2) comparative evaluation across algorithms and scales, (3) Unity ML-Agents integration demonstrating sim-to-sim transfer with successful package delivery, and (4) identification of scaling challenges. Full experimental details and results are documented in our \href{https://pallman14.github.io/MARL-QMIX-Warehouse-Robots/}{Quarto documentation book}.

\section{Related Work}

\textbf{Value Decomposition}: VDN \citep{sunehag2018vdn} factorizes joint Q-values additively ($Q_{tot} = \sum_i Q_i$), while QMIX \citep{rashid2018qmix} uses hypernetwork-based mixing for monotonic relationships. Both follow Centralized Training with Decentralized Execution (CTDE), well-suited for warehouse robotics where deployed robots act on local observations.

\textbf{Independent Learning}: IPPO trains separate policies per agent, treating others as environment. Despite theoretical non-stationarity limitations, it shows surprising effectiveness in cooperative settings \citep{yu2022surprising}. MASAC extends soft actor-critic with centralized critics and entropy regularization.

\textbf{RWARE Environment}: \citet{papoudakis2021benchmarking} provide a standardized warehouse benchmark with sparse rewards, partial observability, and configurable difficulty based on grid size and agent count.

\section{Methods}

\subsection{Problem Formulation}

We model the task as a Dec-POMDP where each agent $i$ receives local observation $o_i$, selects action $a_i \in \{\text{left, right, forward, load/unload, no-op}\}$, and the team receives shared reward based on deliveries. Sparse rewards make credit assignment challenging.

\subsection{QMIX Architecture}

QMIX factorizes the joint action-value function:
\begin{equation}
Q_{tot}(\boldsymbol{\tau}, \boldsymbol{a}) = f_{mix}(Q_1(\tau_1, a_1), ..., Q_n(\tau_n, a_n); s)
\end{equation}
where $f_{mix}$ is a mixing network with non-negative weights ensuring monotonicity ($\frac{\partial Q_{tot}}{\partial Q_i} \geq 0$). We use GRU-based agents (64 hidden units) with 2-layer hypernetwork mixers.

\subsection{Training Configuration}

Our optimized configuration differs substantially from defaults:

\begin{table}[H]
\centering
\small
\caption{Hyperparameter Configuration: Default vs Optimized}
\begin{tabular}{lcc}
\toprule
Parameter & Default & Optimized \\
\midrule
Batch Size & 32 & 256 \\
Buffer Size & 5,000 & 200,000 \\
Epsilon Anneal Time & 50,000 & 5,000,000 \\
Training Steps & 2,000,000 & 20,000,000 \\
Learning Rate & 0.0005 & 0.0005 \\
\bottomrule
\end{tabular}
\end{table}

Extended epsilon annealing proved critical---rapid decay prevents agents from discovering sparse rewards. Larger replay buffers enable learning from diverse experiences.

\subsection{Unity ML-Agents Integration}

We developed a Unity environment mirroring RWARE with 3D visualization (\href{https://pallman14.github.io/MARL-QMIX-Warehouse-Robots/price/week3.html}{implementation details}): 3 agents, 6 discrete actions, 36-dimensional LIDAR-based observations, 200-step episodes. Training uses no-graphics mode with 50x time scaling.

\section{Experiments and Results}

\subsection{Environment Progression}

We validated on MPE before transitioning to RWARE (\href{https://pallman14.github.io/MARL-QMIX-Warehouse-Robots/salmon/week2.html}{transition analysis}). MASAC on MPE converged in 30,000 steps achieving 63\% improvement over random baseline (\href{https://pallman14.github.io/MARL-QMIX-Warehouse-Robots/lian/week1.html}{MPE results}); RWARE required 20M+ steps for comparable performance---a 600x difference in sample complexity. Default hyperparameters produced zero learning after 2M steps, underscoring how dense-reward benchmarks mislead about algorithm readiness for sparse warehouse tasks.

\subsection{Algorithm Comparison}

\begin{table}[H]
\centering
\small
\caption{Performance Comparison on RWARE Environments (Test Return)}
\begin{tabular}{lccc}
\toprule
Algorithm & tiny-2ag-v2 & small-4ag-v1 & Training Steps \\
\midrule
QMIX & \textbf{3.25} & \textbf{2.10} & 20-30M \\
IPPO (Advanced) & 0.38 & --- & 5M \\
IPPO (Vanilla) & 0.13 & --- & 20M \\
Random Baseline & 0.05 & 0.02 & N/A \\
\bottomrule
\end{tabular}
\end{table}

QMIX (3.25) outperformed advanced IPPO (0.38) by 8.5x on tiny-2ag-v2 (\href{https://pallman14.github.io/MARL-QMIX-Warehouse-Robots/price/week2.html}{IPPO analysis}). Even vanilla IPPO with 4x longer training achieved only 0.13---25x lower than QMIX. Value decomposition provides substantial advantages for sparse-reward coordination where agents must learn to implicitly divide labor for package delivery (\href{https://pallman14.github.io/MARL-QMIX-Warehouse-Robots/lian/week4.html}{QMIX learning curves}).

\subsection{Scaling Analysis}

Performance degrades with agent count while training requirements grow super-linearly:

\begin{table}[H]
\centering
\small
\caption{QMIX Scaling Results Across Environment Sizes}
\begin{tabular}{lccc}
\toprule
Configuration & Agents & Test Return & Required Steps \\
\midrule
tiny-2ag-v2 & 2 & 3.25 & 20M \\
small-4ag-v1 & 4 & 2.10 & 30M \\
medium-6ag-v1 & 6 & 1.45 & 40M \\
\bottomrule
\end{tabular}
\end{table}

Scaling from 2 to 6 agents requires 2x more training (20M to 40M steps) while performance drops 55\%. The joint action space grows as $|A|^n$: 6 agents yield 15,625 joint actions versus 25 for 2 agents, suggesting hierarchical approaches may be needed at scale.

\subsection{Unity Deployment Results}

We validated QMIX on Windows Server 2022 (Intel Xeon E5-2680, 196GB RAM):

\begin{table}[H]
\centering
\small
\caption{Unity Training Results on Windows Server}
\begin{tabular}{lcc}
\toprule
Metric & 500K Steps & 1M Steps \\
\midrule
Test Return Mean & 95.2 & 238.6 \\
Peak Return & $\sim$150 & 443 \\
Episode Length & 200 & 200 \\
Training Time & 3h 2min & 6h 18min \\
\bottomrule
\end{tabular}
\end{table}

Console logs confirmed consistent package delivery. At 500K steps, agents showed functional but suboptimal behavior; by 1M steps, policies became nearly deterministic (return std $<$0.01) with smooth navigation. Despite continuous physics challenges versus grid-based RWARE, core coordination behaviors transferred successfully.

\section{Discussion}

Our primary research goal was to investigate the complex task of multiple agents learning to communicate implicitly and coordinate package delivery---a fundamental challenge in warehouse automation. QMIX's value decomposition approach proved effective for this coordination problem, enabling agents to learn complementary roles without explicit communication channels.

\textbf{Key Findings}: (1) Default configurations fail---extended epsilon annealing (5M+ steps) is essential for sparse reward discovery (\href{https://pallman14.github.io/MARL-QMIX-Warehouse-Robots/dre/week2.html}{hyperparameter analysis}); (2) QMIX substantially outperforms independent learning (8.5x on RWARE), validating value decomposition for credit assignment (\href{https://pallman14.github.io/MARL-QMIX-Warehouse-Robots/lian/week4.html}{algorithm comparison}); (3) scaling challenges remain significant---industrial deployments (50+ robots) will require hierarchical approaches (\href{https://pallman14.github.io/MARL-QMIX-Warehouse-Robots/salmon/week3.html}{scaling analysis}).

\textbf{Unity Deployment}: Our Unity experiments demonstrated that learned coordination transfers to 3D environments with continuous physics. On Windows Server, agents achieved 238.6 mean test return after 1M steps, with console logs confirming consistent package pickup and delivery behavior (\href{https://pallman14.github.io/MARL-QMIX-Warehouse-Robots/salmon/week5.html}{deployment results}). The transition from grid-based RWARE to continuous Unity revealed that core multi-agent coordination---implicit task allocation, collision avoidance, and sequential delivery strategies---generalizes across environment representations.

\textbf{Practitioner Recommendations}: Extend epsilon annealing to 5M+ steps (100x default), use large replay buffers (200K vs 5K default), and plan for 20M+ training steps. Recommended configuration: batch size 256, buffer 200K, learning rate 0.0005, GRU dimension 64 (\href{https://pallman14.github.io/MARL-QMIX-Warehouse-Robots/dre/week2.html}{configuration details}).

\textbf{Limitations}: Simulation-only evaluation (no physical robots), testing limited to 2-6 agents, simplified RWARE task structure, and extended training requirements may be impractical for rapid deployment.

\section{Further Research}

Several directions could advance warehouse MARL systems:

\textbf{Hyperparameter Optimization}: Systematic exploration of buffer sizes (128--512 episodes), learning rate schedules, and network architectures. Our 32-episode buffer was relatively small; larger buffers may improve sample efficiency and stability. Automated hyperparameter search could accelerate configuration for new warehouse layouts.

\textbf{Environment Scaling}: Extending beyond 4-6 agents to industrial scales (50+ robots) requires hierarchical decomposition or task partitioning. Larger warehouse grids, dynamic obstacles, multi-floor layouts, and time-sensitive deliveries would better reflect real-world complexity.

\textbf{Algorithm Comparisons}: Benchmarking QPLEX, MAPPO, and MAVEN against QMIX on warehouse tasks would clarify when value decomposition provides advantages versus policy gradient methods. Communication-based approaches may help with larger agent teams.

\textbf{Sim-to-Real Transfer}: Bridging simulation success to physical deployment remains the critical gap. Domain randomization, robust perception, and safety constraints need investigation.

\section{Conclusion}

We demonstrated that QMIX substantially outperforms independent learning for sparse-reward warehouse coordination (8.5x improvement), but requires extensive hyperparameter tuning---particularly extended exploration schedules. Our Unity integration validates sim-to-sim transfer. Scaling to industrial deployments remains an open challenge requiring hierarchical approaches.

\subsection*{Code and Data Availability}

All code, trained models, and experimental results are available at:
\begin{itemize}
    \item GitHub: \url{https://github.com/pallman14/MARL-QMIX-Warehouse-Robots}
    \item Documentation: \url{https://pallman14.github.io/MARL-QMIX-Warehouse-Robots/}
\end{itemize}

\subsection*{AI Assistance Acknowledgment} This research utilized Large Language Models (Claude, ChatGPT) for writing assistance, code debugging guidance, and documentation preparation. All experimental design, algorithm selection, and research conclusions were made by the authors. Comprehensive AI usage details: \url{https://github.com/pallman14/MARL-QMIX-Warehouse-Robots/blob/main/AI_ASSISTANCE_DISCLOSURE.md}
\bibliographystyle{plainnat}

\begin{thebibliography}{5}

\bibitem[Rashid et al.(2018)]{rashid2018qmix}
Rashid, T., Samvelyan, M., Schroeder, C., Farquhar, G., Foerster, J., and Whiteson, S.
\newblock QMIX: Monotonic Value Function Factorisation for Deep Multi-Agent Reinforcement Learning.
\newblock \emph{International Conference on Machine Learning}, 2018.

\bibitem[Sunehag et al.(2018)]{sunehag2018vdn}
Sunehag, P., Lever, G., Gruslys, A., et al.
\newblock Value-Decomposition Networks for Cooperative Multi-Agent Learning.
\newblock \emph{International Conference on Autonomous Agents and Multiagent Systems}, 2018.

\bibitem[Papoudakis et al.(2021)]{papoudakis2021benchmarking}
Papoudakis, G., Christianos, F., Sch{\"a}fer, L., and Albrecht, S.V.
\newblock Benchmarking Multi-Agent Deep Reinforcement Learning Algorithms in Cooperative Tasks.
\newblock \emph{Neural Information Processing Systems Track on Datasets and Benchmarks}, 2021.

\bibitem[Yu et al.(2022)]{yu2022surprising}
Yu, C., Velu, A., Vinitsky, E., Gao, J., Wang, Y., Baez, A., and Wu, Y.
\newblock The Surprising Effectiveness of PPO in Cooperative Multi-Agent Games.
\newblock \emph{Neural Information Processing Systems}, 2022.

\end{thebibliography}

\end{document}